\documentclass{article}

\PassOptionsToPackage{numbers, compress}{natbib}

\usepackage[preprint]{neurips_2023}

\usepackage[utf8]{inputenc} %
\usepackage[T1]{fontenc} %
\usepackage{url} %
\usepackage{booktabs} %
\usepackage{amsfonts} %
\usepackage{nicefrac} %
\usepackage{microtype} %
\usepackage[table]{xcolor} %
\usepackage{natbib}

\usepackage{multirow, multicol}
\usepackage{times}
\usepackage{epsfig}
\usepackage{graphicx}
\usepackage{amsmath}
\usepackage{amssymb}
\usepackage{soul}
\usepackage{pifont}
\usepackage{subfigure}
\usepackage{overpic}
\usepackage{algorithm}
\usepackage{algpseudocode}
\usepackage{wrapfig}
\usepackage{hyperref}
\usepackage{tikz}
\usepackage{pgfplots}
\pgfplotsset{compat=newest}  
\usetikzlibrary{external}
\usepackage{caption}
\usepackage{drawstack}
\usetikzlibrary{matrix,calc}
\newcommand{\ie}{\textit{i.e.}}
\newcommand{\eg}{\textit{e.g.}}

\newcommand{\etal}{\textit{et. al}}
 
\title{Every Mistake Counts in Assembly}

\author{
 Guodong Ding $^1$ %
 \quad
 Fadime Sener $^2$ \quad
 Shugao Ma $^2$ \quad
 Angela Yao $^1$ \\
 $^1$National University of Singapore \quad $^2$Meta Reality Labs Research \\
 {\tt \small \{dinggd, ayao\}@comp.nus.edu.sg \quad \{famesener, shugao\}@meta.com}
}

\begin{document}
\algnewcommand\algorithmicswitch{\textbf{switch}}
\algnewcommand\algorithmiccase{\textbf{case}}
\algnewcommand\algorithmicassert{\texttt{assert}}
\algnewcommand\Assert[1]{\State \algorithmicassert(#1)}%
\algdef{SE}[SWITCH]{Switch}{EndSwitch}[1]{\algorithmicswitch\ #1\ \algorithmicdo}{\algorithmicend\ \algorithmicswitch}%
\algdef{SE}[CASE]{Case}{EndCase}[1]{\algorithmiccase\ #1}{\algorithmicend\ \algorithmiccase}%
\algtext*{EndSwitch}%
\algtext*{EndCase}%
\algnewcommand{\algorithmicand}{\textbf{ and }}
\algnewcommand{\algorithmicor}{\textbf{ or }}
\algnewcommand{\OR}{\algorithmicor}
\algnewcommand{\AND}{\algorithmicand}
\algnewcommand{\var}{\texttt}

\maketitle

\begin{abstract}
One promising use case of AI assistants is to help with complex procedures like cooking, home repair, and assembly tasks. Can we teach the assistant to interject after the user makes a mistake? This paper targets the problem of identifying ordering mistakes in assembly procedures. We propose a system that can detect ordering mistakes by utilizing a learned knowledge base. Our framework constructs a knowledge base with spatial and temporal beliefs based on observed mistakes. Spatial beliefs depict the topological relationship of the assembling components, while temporal beliefs aggregate prerequisite actions as ordering constraints. With an episodic memory design, our algorithm can dynamically update and construct the belief sets as more actions are observed, all in an online fashion. We demonstrate experimentally that our inferred spatial and temporal beliefs are capable of identifying incorrect orderings in real-world action sequences. To construct the spatial beliefs, we collect a new set of coarse-level action annotations for Assembly101 based on the positioning of the toy parts. Finally, we demonstrate the superior performance of our belief inference algorithm in detecting ordering mistakes on the Assembly101 dataset. 
\end{abstract}

\section{Introduction}\label{sec:intro}
We all know the pains of assembling furniture\footnote{A quick online search leads to dozens of articles with titles like \href{https://www.buzzfeed.com/hbraga/easy-to-build-furniture}{``31 Pieces of Furniture You Won't Have a Hard Time Assembling''} 
and \href{https://www.wsj.com/articles/the-secret-to-assembling-ikea-furniture-without-losing-your-sanity-1469557654}{``The Secret to Assembling IKEA Furniture Without Losing Your Sanity''}}, not to mention making mistakes during the procedure.
Imagine now an AI assistant to support complex procedural activities like furniture assembly, cooking, or home repair. An intelligent assistant should be able to learn from the mistakes and detect them in the future. %
Mistake detection can be considered a procedural activity understanding task.  %
Procedural activities are well explored in video understanding, and typical tasks include temporal action segmentation~\cite{ding2022temporal,farha2019ms}, action anticipation~\cite{abu2018will,sener2019zero,sener2022transferring}. %
Making mistakes is a natural and common part of performing procedures in real-world settings. In the video dataset Assembly101~\cite{sener2022assembly101}, adult participants were asked to assemble and disassemble a children's toy vehicle designed for 4- to 6-year-olds, however, nearly $60\%$ of the sequences contained at least one mistake.  $27.4\%$ of these were action ordered incorrectly. For example, as shown in Fig.~\ref{fig:transitive},  if the roof is placed on the cabin before the speaker and the light, it will be impossible to position them afterwards. In developing an AI assistant, it is natural to consider support in the form of assessment and mistake detection.

As a procedural activity, toy assembly distinguishes itself from similar activities in a number of ways. Firstly, assembling a toy involves a fixed set of steps based on the predetermined structure of the parts. Secondly, toy assembly is a sequential process, where each step builds upon the previous one. It is therefore necessary to follow some ordering of actions to complete the toy. Lastly, assemblers may encounter situations requiring problem-solving skills, such as identifying mistakes, locating alternative solutions, and resolving problems. 
The closest related tasks to the notion of mistakes in the video domain are anomaly detection~\cite{sultani2018real} and unintentional action detection~\cite{epstein2020oops}.  These tasks share the same spirit of detecting deviations from expected behaviour. For example, a car accident is an anomaly, whereas knocking over a vase is unintentional. However, it is important to note that anomalies or unintentional actions are inherently defined by their semantics, \ie, observing the acts stand-alone is sufficient to determine that they are atypical. 
The same cannot be said for ordering mistake actions in assembly sequences because a given action can be correct or incorrect depending on the temporal context in which it is being performed, \eg, it is physically possible to attach the cabin before or after attaching the seats. However, `attaching cabin' will only be a correct action when seats are already placed. %

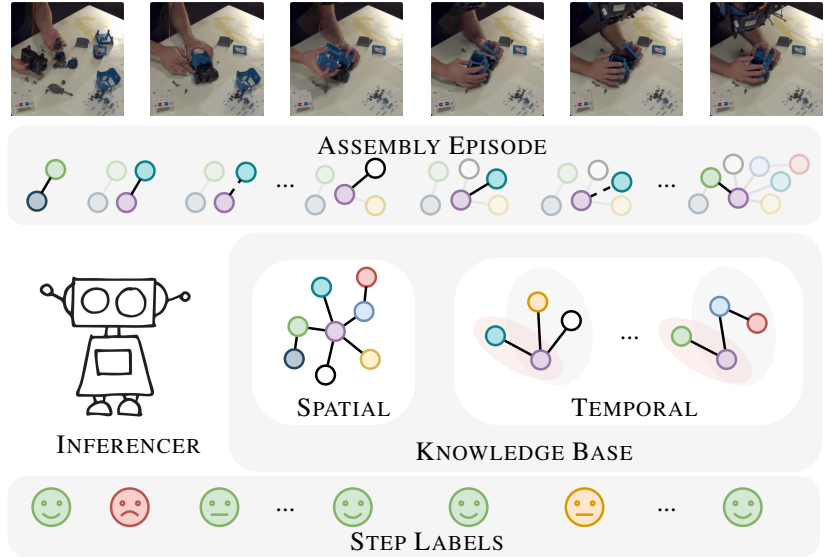
\begin{figure}[t]
    \centering
    \begin{overpic}[width=0.78\textwidth]{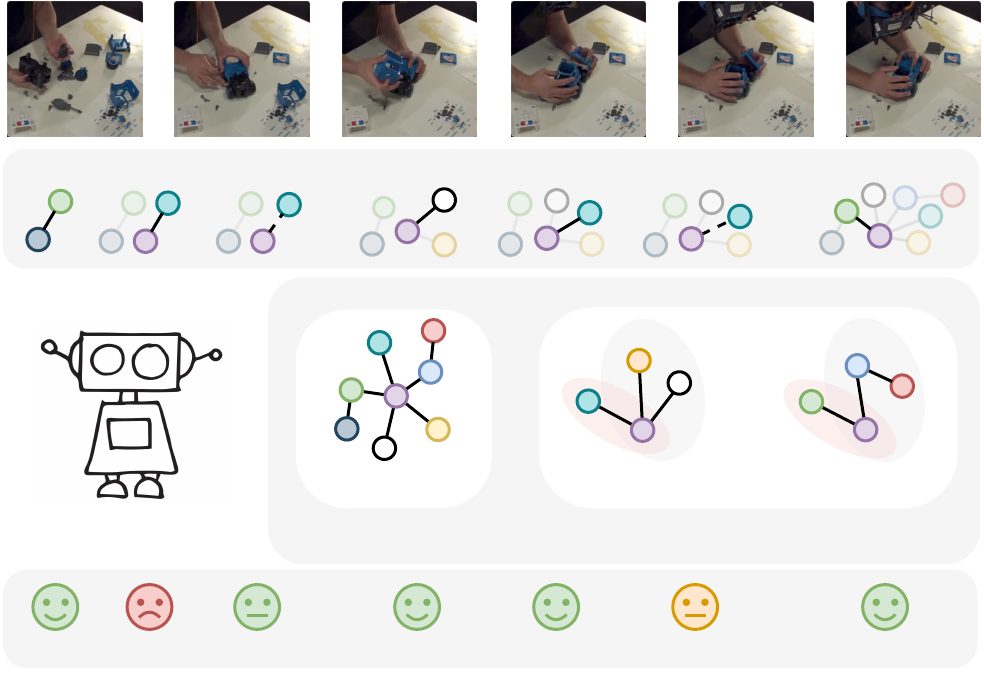}
        \put(6.2,13.5){\textsc{Inferencer}}
        \put(75,27.5){...}
        \put(50,12.7){\textsc{Knowledge Base}}
        \put(35.5,18){\textsc{Spatial}}
        \put(69,18){\textsc{Temporal}}
        \put(79.5,6.5){...}
        \put(33,6.5){...}
        \put(79.5,46){...}
        \put(33,46){...}
        \put(42,1.7){\textsc{Step Labels}}
        \put(38,50){\textsc{Assembly Episode}}
    \end{overpic}
    \caption{Overview of our mistake detection system. Each colored circle denotes a toy part to be assembled. The \textsc{Inferencer} takes as input an assembly sequence incrementally, consults the Knowledge base at each step to make predictions on the action of interest. Faces in the sequence they appear denote `correct',`order mistake',`correction' and `unnecessary detach'. The knowledge base summarizes two type of belief sets, spatial and temporal. Spatial beliefs reveals the topology of objects while temporal beliefs denotes the ordering constraints.
    }
    \label{fig:teaser}
\end{figure}

Knowledge representation and reasoning (KRR) is a fundamental area of study in artificial intelligence (AI). 
Knowledge representation models explicit and implicit knowledge in a domain, ranging from factual data to complex relationships and rules. Reasoning, on the other hand, refers to the ability of AI systems to derive new information or make intelligent inferences based on available knowledge. In this paper, our goal is to develop an intelligent system that can detect ordering mistakes in assembly sequences in the presence of a knowledge base, as depicted in Fig.~\ref{fig:teaser}. 
Specifically, two types of beliefs are formed in the knowledge base, \ie, spatial beliefs and temporal beliefs. 
We define spatial beliefs to accumulate the topological relationships of the toy parts. For example, the `wheel' attaches to the `chassis'; attaching `wheel' to `cabin' is not allowed and should therefore be identified as a mistake. Spatial beliefs are agnostic of the action order, \eg, `wheel' and `cabin' are both attached to the `chassis, but the order in which they are attached is not implied. We therefore introduce temporal beliefs to keep track of the ordering constraints in the action sequences. %

To the best of our knowledge, Sener~\etal~\cite{sener2022assembly101} are the first to advocate for the mistake detection task and their Assembly101 is the only real-world dataset with mistake annotations in action orderings. However, the annotated actions 
are ambiguous, as they reference only a single component. 
An assembly action, however, involves at least two working objects, and the other object must be provided to 
reveal the structural information.
To this end, we provide a new annotations set that includes both interacting parts. 
 
To summarize, our main contributions are threefold: \textbf{1)}
We formulate the ordering mistake detection problem as a knowledge representation and reasoning problem. To this end, we propose a novel mistake detection framework consisting of a \textsc{BeliefBuilder} and an \textsc{Inferencer} to work with an assembly knowledge base. The \textsc{BeliefBuilder} constructs the knowledge base in an online fashion and the \textsc{Inferencer} consults the knowledge base when making predictions. 
\textbf{2)}
We design two types of beliefs in the knowledge base, \ie, spatial and temporal beliefs, both in the format of logic rules. Spatial beliefs define the toy topology, while temporal beliefs encompass the ordering constraints. Additionally, we offer a graph interpretation for each type of belief. 
\textbf{3)}
We enrich the Assembly101 dataset with information about the mistake type and the explicit part-to-part connection details to facilitate the mistake detection task. We evaluate our method on Assembly101 and demonstrate its superior capability for mistake detection.
 
\section{Related Work}\label{sec:literature} 

\paragraph{Procedural Activity Understanding.} Procedural knowledge is an important aspect of cognitive psychology and educational research. It has been studied in computer vision under the context of procedural planning~\cite{chang2020procedure}, step forecasting~\cite{sener2019zero,lin2022learning} and temporal action segmentation~\cite{ding2022temporal}. Common investigated procedural activities by the research community are from the cooking~\cite{Damen2022RESCALING,malmaud2015s,zhou2018towards,zhukov2019cross} or assembly~\cite{sener2022assembly101,ragusa2021meccano,ben2021ikea}. Procedural knowledge can be classified according to how they are sourced, i.e., explicit and implicit knowledge. Explicit knowledge assumes external providers, e.g., in the form of recipes~\cite{salvador2017learning} or instruction manuals~\cite{koupaee2018wikihow}, while implicit knowledge indicates learning from data with no supervision~\cite{sener2019zero,sener2022transferring} or distant supervision~\cite{lin2022learning}.  Our approach falls into the second category as we learn to construct our beliefs from the training data.

\paragraph{Mistake Detection.} 
Our approach is the first to study ordering mistakes in procedural activities that can handle the flexibility of action order and the variation in approaches of the participants.
\cite{Soran2015GeneratingNF}  tried to detect  
missing actions  for making lattes. In their dataset, 18 of the 41 videos  has a purposefully omitted action, \eg \emph{`steaming milk'}. \cite{Soran2015GeneratingNF} model the dependencies of latte-making actions with a directed graph and learn the graph from the complete sequences. Missing actions, however, are not identified until the entire sequence is completed. 
This method is not generalizable to detect the assembly ordering mistakes in Assembly101 since it can only identify the missing steps in a fixed order.

\textbf{Anomaly Detection.}
Detecting anomalies, especially in temporal sequences and unintentional actions~\cite{epstein2020oops,sultani2018real} are tasks similar to ours in spirit. 
These differ from procedural mistakes because the unintended actions are identifiable by their inherent semantics. An example would be a person walking suddenly falling to the ground. The unintionality is defined by the semantics of falling to the ground. On the contrary, the mistakes in an assembly task are usually irrelevant to semantics and more dependent on the temporal context it locates. This makes mistake detection in assembly tasks a suitable case for modeling the temporal logic of action sequences.

\section{The Approach}\label{sec:approach}

\subsection{Preliminaries}\label{sec:annotation}

\textbf{Study on mistakes.}
We first survey all the mistakes made by participants in Assembly101 and list them in Table~\ref{tab:labels}.
Coarsely speaking, there are three classes for the mistake detection task: \emph{`correct'}, \emph{`mistake'} and \emph{`correction'}, where the latter is a step made to rectify the mistake. The mistakes can be further classified into four types according to their root cause. The most straightforward type is the generic ordering mistake. Accumulated mistakes happen after generic order mistakes. Misorientation mistakes 
occur to the placement of 
a part in an incorrect orientation, \eg, a reversed cabin. The last type of mistake is the unnecessary detachment of correctly assembled parts. 
Detecting misorientation mistakes requires 3D perception and a model of the toy parts, which extends beyond the scope of this work. We aim to build a system to identify assembly sequences' ordering mistakes.  %
\begin{table}[tbh]
\centering
\caption{Six types of mistakes in Assembly101. %
Misorientation shown in grey as it is beyond the scope of our work. %
}\label{tab:labels}
\begin{tabular}{ccclcc}
\toprule
Verb                    & Coarse         &   \# of samples         & Remark                   & Fine   &   \# of samples  \\ \midrule
\multirow{4}{*}{attach} & correct         &    2914         & correct step          &     A  & 2914\\ \cmidrule(l){2-6} 
                        & \multirow{3}{*}{mistake}&  \multirow{3}{*}{355} & generic order             &     B     & 153\\ \cmidrule(l){4-6} 
                        &                &            & accumulated  & C&46\\ \cmidrule(l){4-6} 
                        &                 &           &  \textcolor{gray!40}{misorientation} & D&156\\\midrule
\multirow{2}{*}{detach} & mistake           &   371       & unnecessary       & F &371\\\cmidrule(l){2-6}  
                        & correction       &   348         & correction              &         E&348\\ \bottomrule
\end{tabular}
\end{table}

\textbf{Definitions.}
Consider a collection of $N$ assembly sequences $\mathbf{S} =\{\mathbf{s}_n\}_{n=1}^N$ and its corresponding label set $Y$. Each sequence $\mathbf{s} = \{(v, i, j)_t\}_{t=1}^T$ has $T$ steps, where $v\in\{\text{attach}, \text{detach}\}$ denotes the `verb', 
while $(i,j)$ are toy part indices. The step-wise mistake label is denoted as $y_t$. The $(i,j)$ part notations are generally considered commutable in our work, \ie, $(i,j) \equiv (j,i)$. 

The objective of the mistake detection system is to learn from the existing mistakes in action sequences such that it can be used to detect mistakes in future unseen sequences.  %
To do so, we build a knowledge base $\mathcal{K} = \{\mathcal{S},\mathcal{T}\}$ that maintains 
spatial beliefs $\mathcal{S}$ and temporal beliefs $\mathcal{T}$. 
We propose a \textsc{BeliefBuilder} and an \textsc{Inferencer} to interact with the knowledge base accordingly.
Our\textsc{BeliefBuilder} algorithm are carefully designed to deal with streaming action sequences. As mentioned, the ordering mistake actions are context-dependent. Temporal context is important in both belief building and inference stages. To that end, we define an episodic memory $M$ to form a collective temporal context for the algorithms.  %

\subsection{Spatial Beliefs $\mathcal{S}$.} \label{subsec:spatial}
\begin{wrapfigure}{r}{0.5\textwidth}
    \centering
    \begin{overpic}[width=0.45\textwidth]{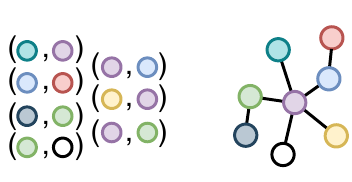}
        \put(0,48){\textsc{Spatial Beliefs:}}
        \put(7,42){$i$}
        \put(16,42){$j$}

        \put(31,38){$i$}
        \put(42,38){$j$}
        
        \put(56,48){\textsc{Rule Graph:}}
    \end{overpic}
    \vspace{-0.5em}
    \caption{Spatial beliefs as a set of part pairs (left) and a rule graph (right).}
    \label{fig:spatial}
\end{wrapfigure}

In the context of toys, the spatial topology is pre-defined, \eg, the `roof' is attached to the `cabin' and the `wheels' to the `chassis'.
Following this logic, we define a spatial belief set, $\mathcal{S}$, which encompasses pairs of connecting toy parts, see Fig.~\ref{fig:spatial} ${(i,j)}$ pairs. The spatial beliefs serve two purposes. To start with, they can be employed to confirm the feasibility of attaching the parts, $i$ and $j$, with the following rule: 

$ \textsc{Spatial}(i,j) \mapsto \hat{y}:$
\begin{align}
 (i,j) \in \mathcal{S} \iff A_{ij}.\label{eq:spatial}
\end{align}

In practice, if any pair is incompatible with the spatial beliefs, attempting to attach them would be 
a mistake, as they would not fit geometrically.  
In addition, the belief set, $\mathcal{S}$, can verify the completion of the assembly sequence and indicate any unperformed or missing actions. %

$\textsc{Completed}(M): $
\begin{align}
         \forall_{(i,j) \in \mathcal{S}} A_{ij}  \in M \iff \textsc{True}
\end{align}

\textbf{Graph Interpretation.} One can represent the spatial belief set as a graph, where each toy part is depicted as a node, and the edges denote the feasibility of attaching them.   
An edge between two nodes indicates feasibility, and completion is achieved when the graph has been fully traversed by the episodic memory $M$. Fig.~\ref{fig:spatial} visualizes the graph representation of spatial beliefs.

\subsection{Temporal Beliefs $\mathcal{T}$.}\label{subsec:temporal}
\begin{figure}[t]
    \centering
    \subfigure[Transitive]{\label{fig:transitive}
    \begin{overpic}[width=0.48\textwidth]{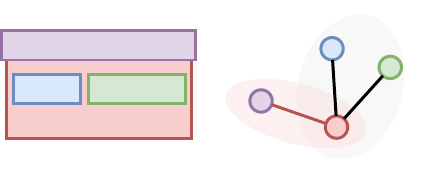}
        \put(-4,40){ \textsc{Part Geometry:}}
        \put(16,11){ \textsc{cabin}}
        \put(2.8,19.5){ \textsc{light}}
        \put(21,19.5){ \textsc{speaker}}
        \put(16,30){ \textsc{roof}}
        \put(47,40){ \textsc{Rule Graph:}}
        \put(58,25){ $r = 1$}
    \end{overpic}
     }
    \subfigure[Intransitive]{\label{fig:intransitive}
    \begin{overpic}[width=0.48\textwidth]{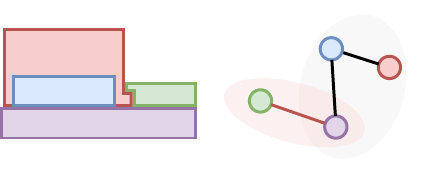}
        \put(-4,40){\textsc{Part Geometry:}}
        \put(14,11){ \textsc{chassis}}
        \put(3,18.5){ \textsc{interior}}
        \put(32,18){ \textsc{base}}
        \put(5.5,30){ \textsc{cabin}}
        \put(47,40){\textsc{Rule Graph:}}
        \put(58,25){ $r= 2$}
    \end{overpic}
    }
    \caption{Transitivity of temporal beliefs. Part geometry implies the ordering constraints (left) and its graph representation (right). Anchor action (the red edge) is reliant on the completion of the rest (back edges in grey group). Transitive and intransitive rules differ in the radius $r$ of their graphs. For transitive rules ($r=1$), any actions from the dependency set (black edges) will be mistakes after anchor action is performed. While, for intransitive rules ($r>1$), only the last action in black edges that completes the rule graph will be considered mistake.} %
    \label{fig:transitivity}
\end{figure}
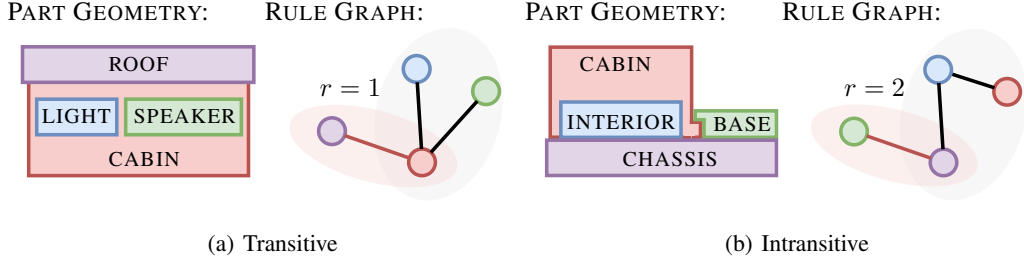

Spatial beliefs do not imply any ordering constraints in attaching the toy parts.
The ordering mistakes are determined by the temporal context in which the action is being observed. 
We determine ordering constraints from the mistake instances during training to establish the temporal beliefs $\mathcal{T}$. 
A temporal belief $T_{ij}$ for its anchor pair $(i,j)$ provides the following rule:

$\textsc{Temporal}(M,i,j) \mapsto \hat{y}:$
\begin{align}
 \forall_{(i',j') \in \mathcal{D}_{ij}} & A_{i'j'} \in M \iff A_{ij}\label{eq:temporal}
\end{align}
where $\mathcal{D}_{ij} \subset \mathcal{T}_{ij}$ corresponds to a list of part pairs $(i',j')$, which we refer to as dependent for the correct execution of $A_{ij}$. In Fig.~\ref{fig:transitive}, with anchor (roof,cabin), $\mathcal{D}_{ij}$ consists of (light,cabin) and (speaker,cabin). The above rule indicates that the anchor action is only possible when all its dependents have been correctly assembled. 

\textbf{Error Accumulation.} %
The \textsc{Temporal} rule considers and predicts only for the anchor action. However, it is necessary to consider the dependents because an error anchor action in context $(\neg A_{ij} \in M)$ would turn its dependent actions from $\mathcal{D}_{ij}$ that are yet to happen into accumulated errors.
To that end, we define the transitivity on $\mathcal{T}_{ij}$ to correspond to two different cases of error accumulation, \ie, \emph{transitive} (written as $\mathcal{T}_{ij}$) and \emph{intransitive} (written as $\neg \mathcal{\mathcal{T}}_{ij}$). A transitive (Fig.~\ref{fig:transitive}) rule indicates that executing any dependent after the anchor is a mistake. With a slight abuse of notation, we define the following inference rule $\mathcal{T}_{ij}(M)$:

$\mathcal{T}_{ij}(M) \mapsto \hat{y}: $
\begin{align}
  \forall_{(i',j') \in \mathcal{D}_{ij}} \exists \neg A_{ij} \in M \to \neg A_{i'j'}
\end{align}

which indicates that with (roof,cabin) attached, attaching (light,cabin) and (speaker,cabin) will be mistakes.
In contrast, an intransitive~(Fig.~\ref{fig:intransitive}) rule would consider executing any dependent to be correct except for the last dependent pair. Consider (base,chassis) attached as the wrong anchor action in context, attaching (cabin,interior) is correct, while further attachment of (interior,chassis) is considered a mistake. The predictions will be the opposite if one first attaches (interior, chassis) and then (cabin,interior), which is implied by the following rule:

$\neg \mathcal{T}_{ij}(M) \mapsto  \hat{y}: $
\begin{align}
     \exists \neg A_{ij} \in M \wedge (\forall_{(i'',j'') \neq (i',j') \in \mathcal{D}_{ij}} A_{i''j''}\in M ) \iff \neg A_{i'j'}\label{eq:intransitive}
\end{align}
We define $\mathcal{T}_{ij}' := \mathcal{T}_{kl} | (i,j) \in \mathcal{D}_{kl} $ the temporal rule that has $(i,j)$ in its dependent set. Combining both transitive and intransitive rule, we have the following to make inference:
\begin{equation}\label{eq:dependent}
    \textsc{Dependent}(M, i, j) \mapsto (\mathcal{T}_{ij}'(M) \wedge \mathcal{T}_{ij}') \vee (\neg \mathcal{T}_{ij}'(M) \wedge \neg \mathcal{T}_{ij}') \\
\end{equation}

\textbf{Graph Interpretation.} Similar to the spatial rule, we show that our temporal rules can be represented as graphs. In its graph form, the transitivity of the rule is determined by its radius. Conventionally, the radius of a graph is defined as:
\begin{equation}
\begin{aligned}
 r = \min_{u\in V}\max_{v\in V} d(u,v)
\end{aligned}
\end{equation}

where $d(u,v)$ is the geodesic distance or shortest-path distance between two nodes $u$ and $v$ in graph $V$. \emph{Transitive} rule graphs have the precise radius with $r=1$ while \emph{intransitive} rule graphs are those with larger radius ($r>1$). An illustration is shown in Fig.~\ref{fig:transitivity}. The anchor action (edge in red) is only correct when the dark edges are traversed. It is possible to consider a hybrid version of these two cases; more details on this case are left to the Supplementary.

\subsection{Belief Building and Inference} \label{subsec:algorithm}
We next focus on the construction and inference procedures of the above belief sets by introducing \textsc{BeliefBuilder} and \textsc{Inferencer}. \textsc{BeliefBuilder} illustrates the process of managing the knowledge base $\mathcal{K}$ to take into account of a new piece of information that is observed incrementally.
As the name suggests, the $\textsc{Inferenecer}$, consults the knowledge base $K$, and makes prediction on the action being observed. 

\begin{algorithm}[H]
\caption{Belief building Step}\label{alg:beliefbuilder}
\begin{algorithmic}[1]
\Procedure{BeliefBuilder}{$M,i,j,y,\mathcal{C}$}
\Switch{$y$}
\Case{$A_{ij}$}
\State $\mathcal{S}\gets \mathcal{S} \cup (i,j)$ 
\State $\mathcal{C}_{ij} \gets  [\mathcal{C}_{ij}]\cap \textsc{Precedes}(M,i,j)$\Comment{Eq.~\ref{eq:context_agg}}
\If{$D_{ij} \in M$}
\State $\mathcal{D}_{ij} \gets [\mathcal{D}_{ij}] \cap [\textsc{Context}(M,i,j)] \cap [\mathcal{C}_{ij}$] \Comment{Eq.~\ref{eq:context_and_candidate}}
\State $\mathcal{T}_{ij} \gets \textsc{Connect}(\mathcal{T}_{ij}, M, \mathcal{S})$ \Comment{Eq.~\ref{eq:connect}}
\State \textsc{pop}($M, \neg A_{ij}$), \textsc{pop}($M,D_{ij}$)
\EndIf
\State \textsc{push}($M, y$)
\EndCase
\Case{$\neg A_{ij}$} 
\If{$\textsc{Accumulated}(\neg A_{ij})$}
\For{$\neg A_{i'j'} \in M, \mathcal{S}$}
\State $\mathcal{D}_{i'j'} \gets \mathcal{D}_{i'j'} \cup (i,j)$ \Comment{Eq.~\ref{eq:acc}}
\EndFor
\EndIf
\State \textsc{push}$(M, y)$
\EndCase
\Case{$\neg D_{ij}$}
\State \textsc{pop}$(M,A_{ij})$
\EndCase
\Case{$D_{ij}$}
\State \textsc{push}$(M,y)$
\EndCase
\EndSwitch
\EndProcedure
\end{algorithmic}
\end{algorithm}

\textbf{\textsc{BeliefBuilder}.}
The knowledge base of spatial beliefs (see Sec.~\ref{subsec:spatial}) and temporal beliefs (see Sec.~\ref{subsec:temporal}) is initialized as empty at the beginning, \ie, $\mathcal{S}=\varnothing, \mathcal{T}=\varnothing$. This indicates that the knowledge base is completely agnostic of the toy being assembled. The \textsc{BeliefBuilder} is aimed to build and update both of them as more assembly sequences are provided in an incremental manner. 
Every mistake counts in assembly sequences, and any ordering mistake always reveals a temporal belief. 
For the mistakes ($\neg A_{ij}$), the mistake context $\textsc{Context}(M,i,j)$ invariably contains its dependents.
The mistake context is the set of correct actions between the mistake fix ($D_{ij}$) and its correct execution ($A_{ij}$), \ie, $\textsc{Context}(M, i,j) = \{ (i',j') |  t_{D_{ij}} < t_{A_{i'j'}}< t_{A_{ij}}, A_{i'j'} \in M \}$. 
$\mathcal{D}_{ij}$ is updated with the following:
\begin{equation}\label{eq:context}
    \mathcal{D}_{ij} \gets [\mathcal{D}_{ij}] \cap [\textsc{Context}(M, i, j)],
\end{equation}
where $[\cdot]$ will only be a valid term if $(\cdot)$ is not an empty set. Although observing the ordering mistakes in the sequences is an obvious and strong cue for the builder to update the temporal beliefs, there is also implicit temporal logic hidden behind a fully correct assembly. For example, an action $A_{ij}$ is not temporally dependent on its subsequent actions. In reverse, its preceding actions $\textsc{Precedes}(M, i,j) = \{(i',j') | A_{i'j'}\in M \}$ would form a candidate set $\mathcal{C}_{ij}$ and any dependents in $\mathcal{T}_{ij}$ fall within that set, \ie, $\mathcal{D}_{ij} \subset \mathcal{C}_{ij}$. In the online setting, we continuously narrow down $\mathcal{C}_{ij}$ with the following:
\begin{equation}\label{eq:context_agg}
    \mathcal{C}_{ij} \gets [\mathcal{C}_{ij}] \cap \textsc{Precedes}(M, i,j)
\end{equation}
Note that the preceding actions depend on $M$ and may change between different episodes, while $\mathcal{C}_{ij}$ are shared across episodes. Taking $\mathcal{C}$ into consideration, Eq.~\ref{eq:context} becomes:
\begin{equation}\label{eq:context_and_candidate}
    \mathcal{D}_{ij} \gets [\mathcal{D}_{ij}] \cap [\textsc{Context}(M, i, j)] \cap [\mathcal{C}_{ij}]
\end{equation}
It is possible (as shown in Fig.~\ref{fig:examplar}) that the algorithm may miss dependents in intransitive temporal beliefs, due to the flexibility of labels for the dependents as discussed in Rule (\ref{eq:intransitive}). Missing dependents cause the intransitive temporal rule graph (Fig~\ref{fig:intransitive}) to be disjoint. To mitigate this situation, we leverage the spatial belief $\mathcal{S}$ to find the minimum number of actions present in the episodic memory $M$ to connect any sub-graphs. Specifically, 
\begin{equation}\label{eq:connect}
    \textsc{Connect}(\mathcal{T}_{ij}, M, \mathcal{S}) \gets \mathcal{D}_{ij} \cup \{(i'j') | (i',j') \in \textsc{Path}(\mathcal{S}, \mathcal{T}_{ij}) \wedge A_{i'j'} \in M\},
\end{equation}
where \textsc{Path}($\mathcal{S},\mathcal{T}$) finds the shortest path in $\mathcal{S}$ that completes the rule graph of $\mathcal{T}$.
To deal with accumulated mistake $\neg A_{ij}$, we would add them into the dependent set of any ongoing order mistakes in context, \ie,
\begin{equation}\label{eq:acc}
    \mathcal{D}_{i'j'} \gets \mathcal{D}_{i'j'} \cup (i,j)  |  \neg A_{i'j'} \in M.
\end{equation}

The \textsc{BeliefBuilder} (Alg.~\ref{alg:beliefbuilder}) is repeated for each step in the episodes. We show the overall algorithm in the Supplementary.

\begin{algorithm}[H]
\caption{Inference Step}
\begin{algorithmic}[1]
\Procedure{Inferencer}{$M, v, i, j$}
\Switch{$v$}
\Case{`attach'}
\State  $\hat{y} \gets $  \textsc{attach}($M,i,j$) \Comment{Eq.~\ref{eq:attach}} %
\State \textsc{push}$(M, y)$
\EndCase
\Case{`detach'}
\State $\hat{y} \gets$ \textsc{detach}($M,i,j$)\Comment{Eq.~\ref{eq:detach}}
\If{$\hat{y} == D_{ij}$}
\State \textsc{pop}($M,\neg A_{ij}$)
\Else
\State \textsc{pop}($M, A_{ij}$)
\EndIf
\EndCase
\EndSwitch
\State \textbf{return} $\hat{y}$
\EndProcedure
\end{algorithmic}
\label{alg:inferencer}
\end{algorithm}

\textbf{\textsc{Inferencer.}} 
The \textsc{Inferencer} is also recurrent and makes predictions based on the belief sets and the previous decisions. Alg.~\ref{alg:inferencer} summarizes the \textsc{Inferencer} step.  At each time step $(M, v, i, j)$,  the \textsc{Inferencer} estimates for the tuple $(v, i, j)$ the mistake label based on the episodic memory $M$, which has historical action labels up to the current step. When the action is to attach, \ie, $v=$`attach', multiple inference rules from both spatial and temporal beliefs, \textsc{Spatial} (Rule.~\ref{eq:spatial}), \textsc{Temporal} (Rule.~\ref{eq:temporal}) and \textsc{Dependent} (Rule.~\ref{eq:dependent}) will be applied jointly to decide on its label $\hat{y}$. While for a detach action that is being perceived at the moment, its label will depend on the episodic memory $M$. If the attachment of the same object pairs exists as a mistake in the memory, this action would be regarded as a `correction', and otherwise it is a mistake of `unnecessary detach'. Overall, the \textsc{Inferencer} makes predictions with the following rule:

$\textsc{Attach}(M, i, j) \mapsto \hat{y}:$
\begin{align}
\textsc{Spatial}(i,j) \wedge \textsc{Temporal}(M,i,j) \wedge \textsc{Dependent}(M,i,j) \iff A_{ij} \label{eq:attach}\
\end{align}
$\textsc{Detach}(M, i, j) \mapsto \hat{y}:$
\begin{align}
    \exists \neg A_{ij} \in M \iff D_{ij} \label{eq:detach}
\end{align}

\section{Dataset and Experiments}\label{sec:experiment}
\subsection{Dataset.} %

Assembly101~\cite{sener2022assembly101} is the first to propose the mistake detection task in assembly sequences. The action labels in the dataset are `(verb, noun)', \eg, `attach arm\_connector'. However, such annotations without mentioning the other interacting part in context introduce inconsistency and ambiguity. For example, two distinct actions attaching `arm\_connector' to `chassis' or `boom' would share the identical action label. Therefore, such annotations are not informative enough for the mistake detection task where spatial connectivity matters.   To that end, for the assembly sequences, we create a new set of part-to-part annotations which includes both interacting objects, denoted in the form of $(\mathtt{verb},\mathtt{this},\mathtt{that})$. Due to the nature of our approach, which focuses on logical reasoning,  we only consider two types of $\mathtt{verb}$s, \ie, $\mathtt{attach}$ and $\mathtt{detach}$. We hope that such explicit part annotations on Assembly101~\cite{sener2022assembly101} encourage more investigation on the task.

\textbf{Self-looped and repetitive actions.} Certain toy parts from Assembly101 can be further split into two halves, \eg, `chassis part' and `water tank part', as illustrated in supplementary. For simplicity, we keep a consistent level of annotation and do not consider the subpart level and annotate them as `attach chassis and chassis'. In this sense, any composition of subparts is a self-looped action. 
Due to the geometric symmetry of the toys, it is common for assembly sequences to involve repetitive steps. For example, four wheels can be attached at different sequential locations; they are annotated whenever they occur, regardless of their number of occurrences.

\textbf{Splits.} There are 328 unique action sequences for assembling 101 toys in the Assembly101 dataset. To create our splits, we randomly sample one action sequence as the test set and use the remaining sequences as the training data for each toy. This process is repeated four times to obtain four different splits, and we report the results as the average over the four splits.

\textbf{Evaluation metric.} We propose the following evaluation metrics to assess the mistake detection performance comprehensively. We report per class recall and precision, while Acc and mean F1 scores are reported over all classes.

\begin{table}[t]
\centering
\caption{Performance comparison with coarse mistake labels. }\label{tab:perf}
\begin{tabular}{lcccccccc}
\toprule
\multirow{2}{*}{} & \multicolumn{2}{c}{mistake} & \multicolumn{2}{c}{correction} & \multicolumn{2}{c}{correct} & \multirow{2}{*}{Acc} & \multirow{2}{*}{F1} \\ \cmidrule(lr){2-3}\cmidrule(lr){4-5}\cmidrule(lr){6-7}
 & recall & precision & recall & precision & recall & precision &  &  \\ \midrule
TempAgg~\cite{sener2020temporal} &36.6& 54.0&49.3&43.4&93.2&79.7&78.8&59.9\\
LSTM & 35.2 & 58.7 & 43.7 & 49.9 & 99.0 & 88.5 & 82.3 & 61.3 \\

Ours & 70.6 & 63.7 & 40.8 & 87.2 & 92.7 & 93.1 & 86.0 & 71.8\\ \midrule
\cellcolor{gray!10}{Gains} & \cellcolor{gray!10}{\textcolor{blue}{+35.4}} & \cellcolor{gray!10}{\textcolor{blue}{+5.0}} & \cellcolor{gray!10}{\textcolor{red}{-2.9}} & \cellcolor{gray!10}{\textcolor{blue}{+37.3}} & \cellcolor{gray!10}{\textcolor{red}{-6.3}}&\cellcolor{gray!10}{\textcolor{blue}{+4.6}}&\cellcolor{gray!10}{\textcolor{blue}{+3.7}}&\cellcolor{gray!10}{\textcolor{blue}{+10.5}}\\
\bottomrule
\end{tabular}
\vspace{-1em}
\end{table}

\subsection{Experiments}

\textbf{Baselines.} Following~\cite{sener2022assembly101}, we adopt the long-range temporal model TempAgg~\cite{sener2020temporal} as the mistake detection model, train with our part-to-part annotation for 15 epochs. In essence, the mistake detection problem can also be modeled using a recurrent neural network (RNNs) by treating it as a sequence-to-sequence task. As another baseline model to compare, we design an LSTM~\cite{hochreiter1997long} comprising four hidden layers, and each hidden layer size is set to 256. We feed the one-hot action feature vector as input for each step and the action sequences are truncated or padded to a fixed length of 60. We train the LSTM on the training data with a learning rate of $1e^{-3}$ for 100 epochs. 

\vspace{-1em}

\paragraph{Coarse mistake detection.} We report the results on Assembly101 dataset trained and evaluated on the coarse mistake labels for different approaches in Table.~\ref{tab:perf}. Both TempAgg~\cite{sener2020temporal} and LSTM take the one-hot encoded feature vector as input. LSTM slightly outperforms TempAgg at the Acc ($+3.5\%$) and F1 ($+1.4\%$) scores. Such a performance gap mainly results from LSTM's boost in `correct' class. Our method achieves a high recall of $70.6\%$ on the mistake class, which doubles that ($35.2\%$) of the LSTM. In the meantime,  ours is by $5\%$ and $9.7\%$ higher in mistake precision than LSTM and TempAgg, respectively. While for both  `correction' and `correct' classes, our approach achieves lower recall values but has a high precision value. This indicates that our approach has a high confidence in the accuracy of instances from these classes. When evaluated across the classes, ours is the best in both Acc ($86.0\%$) and F1 ($71.8\%$), showing its strong ability to capture the ordering dynamics in assembly sequences.

\begin{figure}[h]
\vspace{-1em}
    \centering
    \subfigure[LSTM]{\label{fig:lstm}
\includegraphics{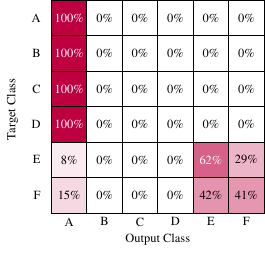}}
    \hspace{1em}
    \subfigure[Ours]{\label{fig:ours}
\includegraphics{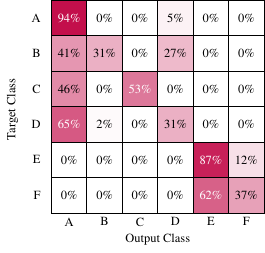}}
\caption{Confusion matrix comparison on the fine mistake labels.}\label{fig:fine}
\vspace{-1em}
\end{figure}

\textbf{Fine mistake detection.} We then compare the performance between our approach and LSTM on the fine mistake labels. The confusion matrix for each approach is plotted in Fig~\ref{fig:fine}. As it is shown in Fig~\ref{fig:lstm}, the the LSTM model practically missed all ordering mistakes (on B, C, and D) and predicted them as correct. This is likely due to the significant imbalance ratio between each fine mistake class and the correct class (see Table.~\ref{tab:labels}). However, the confusion LSTM makes in the bottom right corner indicates the fact that LSTM is picking up the knowledge that a detach action can either be a mistake or a correction. In contrast, our approach (Fig~\ref{fig:ours}) is better at detecting the ordering mistakes and does not confuse a detach action (E and F) to a correct attach class (A) as the LSTM does. 

\begin{figure}[h]
\centering
\scalebox{0.8}{\includegraphics{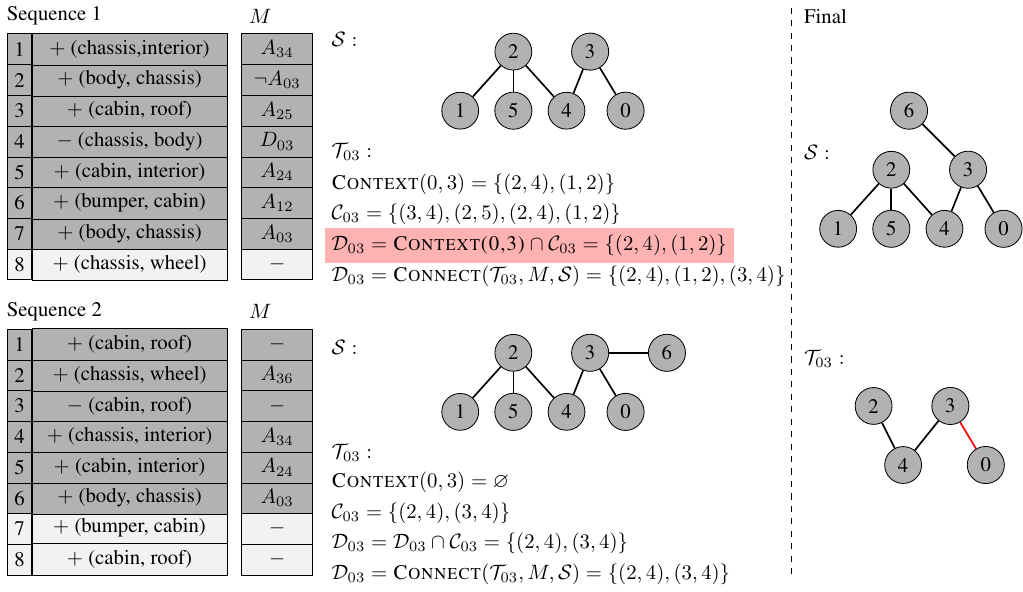}}
\caption{A running example of our \textsc{BeliefBuilder}. Note that at the 7-th step in the first sequence, the $\mathcal{D}_{03}$ before the (highlighted in red) misses one dependent action $(3,4)$ since it is located outside the mistake context and includes an extra pair $(1,2)$ given the mistake context (Steps 5-6). The missing action is retrieved via the \textsc{Connect} step and the extra pair is removed when the \textsc{RuleBuilder} sees the 6-th step in Sequence 2. The right side shows the final spatial and temporal rules.}\label{fig:examplar}
\end{figure}

\textbf{Spatial and temporal beliefs.} We show an example of the beliefs building in Fig.~\ref{fig:examplar}. 
We additionally run our approach on the full set of sequences on Assembly101 dataset and yield in total 48 temporal beliefs. Among these, 23 are transitive, and the remaining 15 are intransitive. It is worth noting that these temporal rules are well aligned with the geometric constraints of the toy parts.

\section{Conclusion}\label{sec:conclusion}
This work is aimed at detecting ordering mistakes in toy assembly sequences. Accordingly, we propose a novel framework that maintains two belief sets to describe the spatial structure and temporal constraints in assembly. The belief sets are constructed in an online fashion with the \textsc{BeliefBuilder} and serve as rules for the \textsc{Inferencer}. Our approach achieves promising ordering mistake detection results and consistently outperforms other approaches on the Assembly101 dataset.

{ 
\bibliographystyle{plain}
\bibliography{references}
}

\end{document}